\documentclass[%
nofootinbib,
amsmath,amssymb,
aps,
showkeys
]{revtex4-2}

\usepackage{graphicx}
\usepackage{dcolumn}
\usepackage{bm}
\usepackage{xspace}
\usepackage{xcolor}
\usepackage{definitions}
\usepackage{caption}
\usepackage{subcaption}

\usepackage{algorithm}
\usepackage{algpseudocode}
\usepackage{silence}
\usepackage{tcolorbox}

\usepackage[mathlines]{lineno}

\begin{document}

\preprint{APS/123-QED}


\title{altiro3D: Scene representation from single image and novel view synthesis}

\author{L. Tenze and E. Canessa}
\email{canessae@ictp.it -- Orcid: 0000-0002-9581-9419}
\affiliation{The Abdus Salam International Centre for Theoretical Physics, ICTP, Trieste 34151, Italy}

\date{\today}

\begin{abstract}
We introduce altiro3D, a free extended C++ library developed to represent reality starting from a given 
original RGB image or flat video. It allows to generate a light-field (or Native) image or video 
and get a realistic 3D experience. To synthesize {\it N}-number of virtual images and add them 
sequentially into a Quilt collage, we apply MiDaS models for the monocular depth estimation, simple 
OpenCV and Telea inpainting techniques to map all pixels, and implement a "Fast" 
algorithm to handle 3D projection camera and scene transformations along {\it N}-viewpoints. We 
use the degree of depth to move proportionally the pixels, assuming the original image to be at 
the center of all the viewpoints. altiro3D can also be used with 
DIBR algorithm to compute intermediate snapshots from a equivalent "Real (slower)" camera 
with {\it N}-geometric viewpoints, which requires to calibrate a priori several intrinsic and 
extrinsic camera parameters. We adopt a pixel- and device-based Lookup Table to optimize 
computing time. The multiple viewpoints and video generated from a single image 
or frame can be displayed in a free-view LCD display.
\end{abstract}


\maketitle


\section{\label{sec:intro}Introduction}

There has been a growing interest in 3D scene representation and multiview synthesis 
using a RGB-D (color and depth) input image or a
pair of stereoscopic images and their relative depth map (see e.g.,~\cite{Ana21}, 
Morpholo~\cite{Can20,Can20a}). These efforts may lead to impressive realistic
results but still incur high computational complexity, making them ill-suited to
run live streaming applications. Alternative methods that have the key advantage of 
rendering the 3D computer reconstruction faster, are those novel methods starting from 
just a single, query RGB image given as the input, see e.g.,~\cite{Eig14,Cha20,Min21,Ran22,Zoe23}.
This new possibility arises with the recent advancements in modeling the 3D visual 
world through the generation of depth maps from monocular images. These can be produced 
by convolutional neural network (CNN) models such as the so-called MiDaS (2.1 and 3.1) algorithms
trained over large-scale RGB datasets (6 and 12, respectively)~\cite{Ran22}. Tremendous 
progress has been done with this alternative framework in recent years~\cite{Pan23,Chau23}.

The purpose of this work is to introduce
altiro3D\footnote{{\it 'altiro'}: Real Academia Española: adv. coloq. Chile. {\it 'inmediately'} (to the point, fast).}, 
a new 2D-to-3D image and video conversion library for free-view LCD designed to 
efficiently compress all the image processing time to approach realistic 3D rendering.
This is a free, extended C++ library developed to reconstruct reality starting from a 
still RGB image or flat video to generate a light-field (or Native) image or video 
and get a realistic 3D experience. In order to synthesize {\it N}-number of virtual images and add
them sequentially into a $N\times M$ Quilt collage, we apply MiDaS coding for the monocular depth 
estimation~\cite{Ran22}, the OpenCV mapping~\cite{OpenCV} and Telea~\cite{Tel04} inpainting techniques
to map all pixels, and implement a "Fast" algorithm to mimic 3D projection 
and scene transformations along the synthesized {\it N}-viewpoints. 
altiro3D uses the degree of depth to move proportionally the pixels, assuming the original image to be 
at the center of all the viewpoints. altiro3D can also be used with the Depth Image Based Rendering
(DIBR) algorithm to compute in-between snapshots from a equivalent "Real (slower)" camera
with {\it N}-geometric viewpoints, which requires to calibrate a priori several intrinsic and
extrinsic camera parameters~\cite{Feh04}. We adopt a pixel- and device-based Lookup Table (LUT)
taking into account display calibration data for specific 3D monitors. As discussed in~\cite{Can20,Can20a}, 
the implementation of LUT allows reduction in the computing time of about
50\%. The multiple viewpoints and video generated from single-shot data
can be displayed in any free-view LCD display~\cite{Tak11}, such as the slanted lenticular Looking
Glass (LG) Portrait~\cite{lgp}. The latter is a low-cost lenticular 3D device which allows to reduce loss of 
resolution in the horizontal direction by slanting the structure of the lenticular lens.

altiro3D is implemented with the goal of minimizing the image processing time to approach real time 
applications in 3D streaming without the need for the viewer to wear any 
special 3D glasses.  On the whole, altiro3D command lines allows to effectively  \\
 \textbullet \; Create Native (i.e., 3D image) from photo using MiDaS small. \\
 \textbullet \; Create Native from photo using MiDaS large. \\
 \textbullet \; Create Native from a given original RGB image and depth image. \\
 \textbullet \; Create Native from a given Quilt (N$\times$M). \\
 \textbullet \; Create Native from sorted {\it N}-views (i.e., sequential set of plain images) stored in a given directory. \\
 \textbullet \; Convert Quilt views to 2D video (.mp4). \\
 \textbullet \; Convert given 2D video to Native 3D video (.mp4). 

Our work leverages the advancement on 3D vision using a single image or frame and presents a
framework designed for the Linux O.S. environment. Moreover, our work does not require the use
of heavy computing runtime, thus can support a wide range of application scenarios in education and science,
among others. The visual quality of our synthesized views provides a rather realistic immersive experience.

\section{\label{sec:related}Related work: Depth map from single image}

As discussed in~\cite{Wan23}, monocular deep estimation networks are classified
into these categories: supervised, unsupervised and self-supervised 
learning. The first to apply supervised learning for monocular image depth 
estimation were Eigen et al.~\cite{Eig14}. Their method uses for training, the input 
image and the corresponding depth map to directly output the depth prediction. 
Semi-supervised methods can obtain the corresponding depth map by training with 
less data sets~\cite{Sha23}. Unsupervised and self-supervised learning methods enables the 
network to perform deep predictions from unlabeled images. Some approaches pass 
the whole image into the network and perform convolution operation to only capture
local information. This limits passing information to other sequence representations 
and leads to low prediction accuracy. Despite these limitations, deep 
learning-based monocular depth estimation that uses, e.g., CNN is a growing 
area of research. These are methods limited to those scenarios present in 
the process of training on the datasets~\cite{Che22,Cha23}.

Recent review articles on the state-of-the-art development
and representative algorithms for deep learning-based monocular depth estimation, which perform 
more accurately under the many restricted conditions, can be found in~\cite{Cha20, Min21}.
They review some mainstream monocular depth estimation methods 
based on deep learning with examples according to diﬀerent datasets training.
In particular, included are a variety of supervised learning methods to address the monocular depth estimation, 
in terms of (i) CNN-based method: to capture depth features layer by layer through their convolution kernels 
and recover depth maps by deconvolution to meet the spatial features of the scene;
(ii) recurrent neural network (RNN): designed to learn spatial–temporal features 
from video sequences; (iii) generative adversarial network (GAN): introduced to generate 
and discriminate between depth maps. The confrontation between a generator and discriminator 
function facilitates the training of the framework. These surveys also introduce publicly 
available datasets and evaluation metrics that have made significant contributions 
to monocular depth estimation.

Very recently, a biologically inspired deep learning network for monocular depth estimation
has been reported in~\cite{Wan23}. This is based on a relationship between the 
self-attention mechanism in biological visual systems and the monocular depth estimation
network. The input to the network are normalized 3D (RGB) pixel values and information 
interaction is established between an encoder, decoder and self-attention fusion unit. The 
function of the encoder is similar to the retina, processing visual information through
integration, and transmitting the information to a next-level module. The information 
transfer between each module in this bio-network mapping enables the deep learning network 
to output a depth map with rich object information and detailed information. 

In our work we follow some of the ideas in~\cite{Cha20,Min21} and apply a CNN-based method.
Specifically we utilize MiDaS models~\cite{Ran22} to process an input image and produce multiple 
views of a scene. By estimating a reasonable accurate depth map in this way, we render 
synthesized views with a DIBR process~\cite{Feh04}. Output pixels in the $N$-views are shifted copies 
of the input image's pixels. Unlike prior work, the depth map generated within our altiro3D algorithm
is not compared against any real depth map and it serves the purpose to represent horizontal 
parallax upon geometric constraints between image sequences. As another difference, computing 
calculations are speed up by implementing a pixel- and device-based LUT as in~\cite{Can20,Can20a}.

\begin{figure}[ht]
    \centering
    \includegraphics[width=0.44\textwidth]{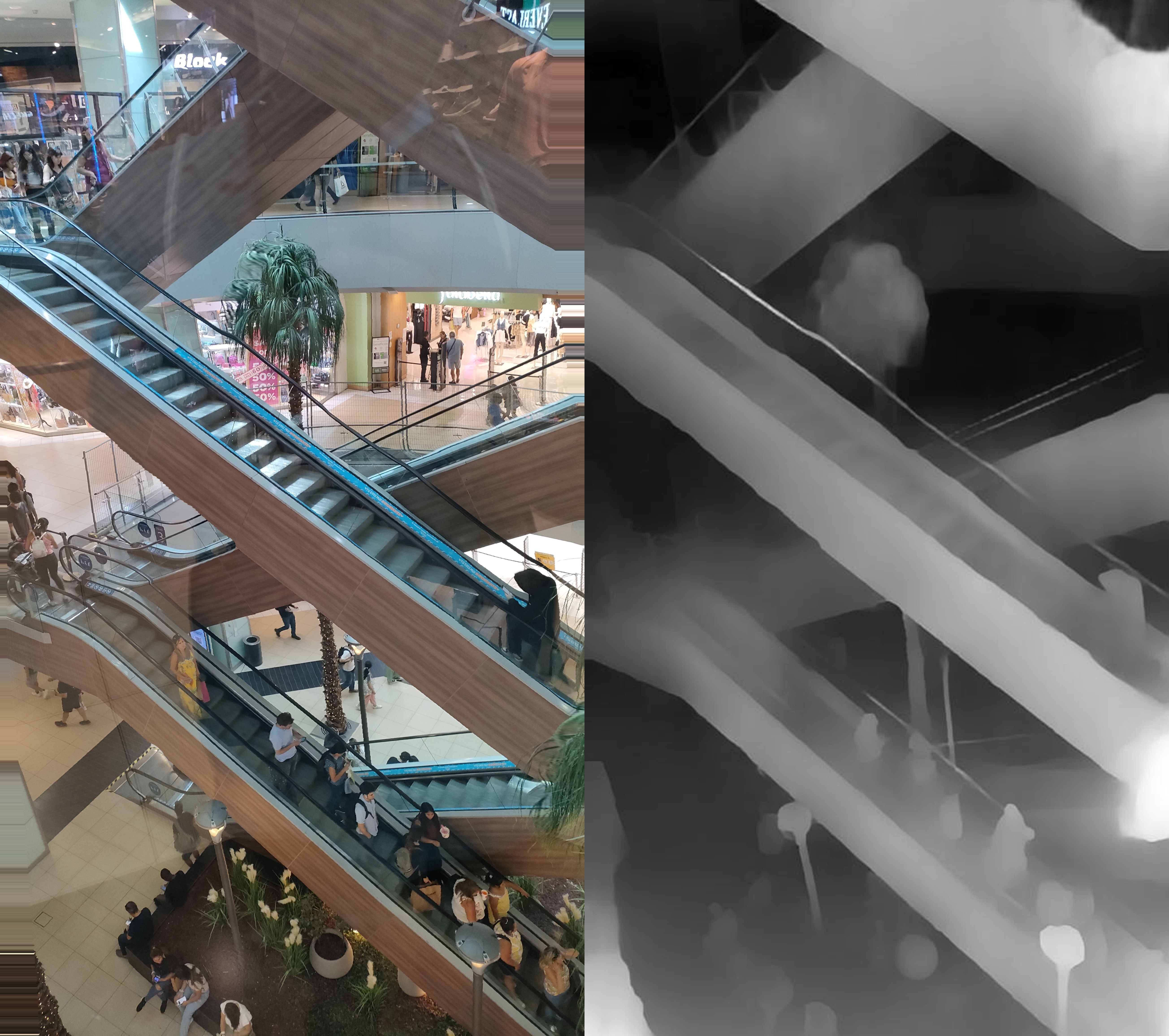} \\
    \includegraphics[width=0.372\textwidth]{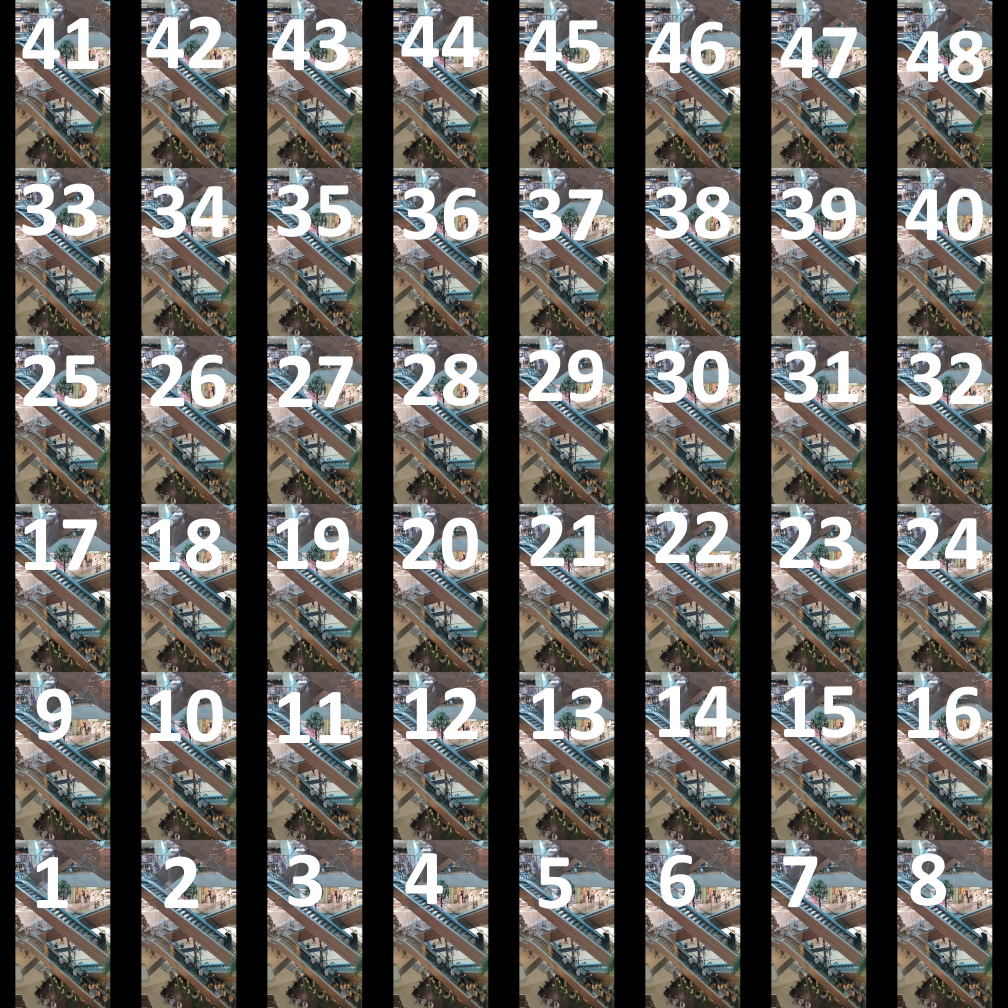}
    \includegraphics[width=0.4\textwidth]{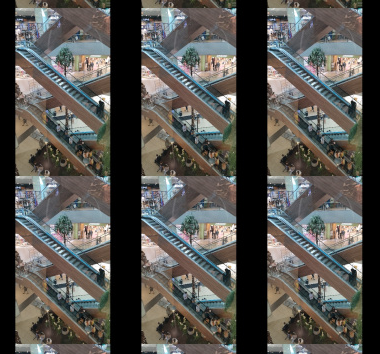}  \\
    \caption{Upper left: Original RGB input photo. Upper right: Depth map obtained using MiDaS~\cite{Ran22}. Lower left: (6x8) Quilt tile containing sequential views of a scene, starting from the bottom-left tile as the input image. Lower right: A section of the Quilt.}
    \label{fig:allimages}
\end{figure}

\section{\label{sec:implementation}Implementation}

The simplest hardware needed to implement altiro3D is a standard PC Computer
(Intel Core i5, 64bit and at least 4G RAM), running a recent release of Linux O.S.
(22.04LT or newer) and any slanted lenticular display such as the LG Portrait \cite{lgp}.
This is an external HDMI video monitor which provides a novel glasses-free
way to preview 3D objects and scenes within an exteded FoV.

The overall system has been developed in C++ and deeply exploits libraries Qt v.5~\cite{QT}
and OpenCV v.4~\cite{OpenCV2}. The system consists in a main library, \verb+libaltiro3D+, where
all developed algorithms have been implemented together with a set of program tools exploiting the
library routines. All important aspects of the library have been properly documented with
Doxygen~\cite{Doxygen} in order to provide other users a good reference to use the library.

The present altiro3D library is an evolution of our previous Morpholo library~\cite{Can20,Can20a}.
These two have been designed taking into account the naming convention used in the literature related to
the field of 3D rendering and in particular, to the conventions used by the display produced by the LG.
So Quilt, Native, Multiviews and other similar terms are used to identify graphical objects and
processing steps inside the new altiro3D library. In addition to the algorithm to produce $N$-views
feeding the holographic display, the Morpholo library implemented both routines to get info about
the calibration of the target display and functions to produce the LUT in order to speed up the mapping
from Quilt to Native.

With respect to the previous Morpholo library, the altiro3D library has been integrated with sources
taking into account the CNN neural network inference and two new models to generate, from a single 
image, intermediate views. These views are used to create the Quilt and, hence, to produce the Native 
image. Both these steps have been optimized to speed up the processing and to approach future real-time
implementations. The CNN processing exploits the capabilities provided by OpenCV to get the inference
from the deep neural network (DNN) and is able to use the CUDA core, if present, or the usual CPU, as
a fallback solution. Some parts of the source code (such as the creation of the views) have 
been optimized by using the parallel architecture provided by OpenCV. The problem image can be
divided in some parts and each part can be processed by a CPU core, improving the speed.

To achieve the goal of a scene representation from single image and novel view synthesis,
altiro3D ({\it i}) takes into account display calibration data for each specific 3D monitor
(including lenticular pitches, slope, screen height and width and number of lens per inch);
({\it ii}) implements a one-time configuration LUT as a simple array indexing operations
that save runtime computation of Eq.(\ref{eq:1}) below; ({\it iii}) uses the MiDaS 2.1 
DNN for a robust monocular depth map estimation. Although the MiDaS 2.1 small model
(included in altiro3D code) cannot provide complete depth information
on distant regions. We mainly extrapolate information from nearby pixels only.
Intermediate views generated using altiro3D are then added into a Quilt collage sequentially
from a given $N\times M$ number of computed intermediate snapshots as shown in the example of Fig.~\ref{fig:allimages}.

The processes involved and repeated within altiro3D (for the creation of any Native 
image/frames output as shown in Fig.~\ref{fig:allimages}), include 

\begin{tcolorbox}
\hspace*{0.1cm}\textbullet \; identify LCD device-related parameters (e.g., the LG Portrait's "visual.json" file) \\
\hspace*{0.4cm}\textbullet \; set input image resolution (e.g., $560\times 420 px$) and Quilt size (e.g., $6\times 8px$)  \\
\hspace*{1cm} $\rightarrow$ generate LUT binary file (e.g., "portrait-6x8.map") \\
\hspace*{3cm}           $\rightarrow$ generate Quilt (e.g, resolution $3360\times 3360 px$) given MiDaS model\\
\hspace*{4cm}                   $\rightarrow$ generate Native: 3D image or 3D video (e.g., size $1536\times 2048 px$) \\
\hspace*{5cm}                           $\rightarrow$ send the Native to display on a free-view LCD device.
\end{tcolorbox}

\subsection{Calibration data from LG portrait}

The device used to test the developed library is the LG Portrait holographic display. In order 
to proper map the pixels from Quilt image to the Native one, it is necessary to acquire the 
calibration of the display. As mentioned, the calibration data changes from device to device, 
so the calibration file is vital to produce the Native image.

The LG Portrait display provides the calibration data available from a simulated storage 
device accessible from the USB port. The user can connect the PC to the display to read the 
internal storage and to get the required file "visual.json".
The internal storage is provided by a Raspberry PI embedded in the Portait display.

\subsection{LUT map from calibration file, Quilt and inpainting}

The LG Portrait per-device calibration file is necessary to create the LUT to properly
create the Native image. The altiro3D suite provides the tool \verb+altiro3Dnative+ to exploit the data
inside the calibration file and to create the LUT. The parameters required to generate the mapping are:
the resolution of each image inside the Quilt and the number of rows and columns of the Quilt. From these
arguments the program produces an output file, usually with extension \verb+.map+, containing
the map matrix, which is crucial for the other altiro3D generation steps and for all the other
provided command tools.

The LUT is created only once at the beginning of the needed device-dependent
mapping. Three matrices are allocated for the color channels RGB. Each matrix provides the X coordinate
of the Quilt from which one takes the corresponding value and the Y coordinate. The multiplication by 2,
allows to avoid unnecessary waste of resources and consume the least possible amount of RAM memory. All
positions of the pixels are considered and one then calculates the mapping value
for each pixel in the Quilt image. This value is stored in the 3 different allocated matrices and each element of the
matrices is made of type uint16\_t. The matrices are then saved in binary format and these are reload
(without recalculating) when applying the mapping to all the $N\times M$ images. This LUT procedure allows
to speed up significantly the needed mapping procedure --the rendering operation of the final native image
which is essentially achieved by accessing the elements of the 3 matrices to map the Quilt pixels.
The resolution of the LG display system corresponds to the $1536\times 2048$ pixels.
Each of the LG Portrait devices combine light-field
and volumetric technologies, and have specific display calibration values for a correct rendering.
This class of lenticular, autostereoscopic display require multiple views of a scene to
provide motion parallax and get a realistic 3D experience by perceiving different stereoscopic pairs.

A Quilt allows for an efficient way to store $N\times M$ images, or frames from a video,
forming a collage ordered as shown in Fig.~\ref{fig:allimages} Quilts serve to save disk space and fast retrieving
images to be displayed. 
The $N\times M$ images forming the Quilt, are converted into a light-field image via the following
expression for the relation between the pixels of a slanted lenticular 3D LCD and the multiple perspective
views~\cite{Ber97}
\begin{equation}
N_{i,j} = N_{tot} (i - i_{_{off}} - 3_{j} tan(\alpha)) mod(P_{x})/P \;\;\; ,  \label{eq:1}
\end{equation}
where $i$ and $j$ denote the panel coordinates for each sub-pixel. Each sub-pixel on
the 3D LCD is mapped to a certain view number and color value (i.e., in the light-field domain).
$N$ denotes the view number of a certain viewpoint, $\alpha$ the slanted angle between the lenticular lens
and the free-view LCD panel and $P_x$ the lenticular pitch.

\begin{figure}[ht]
    \centering
    \begin{subfigure}[b]{0.3\textwidth}
        \includegraphics[width=\textwidth]{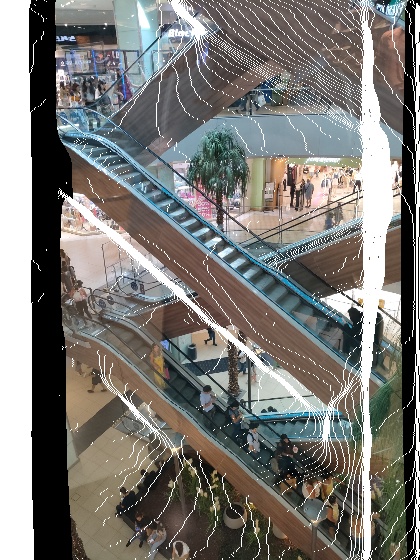}
    \end{subfigure}
    \hfill
    \begin{subfigure}[b]{0.3\textwidth}
        \includegraphics[width=\textwidth]{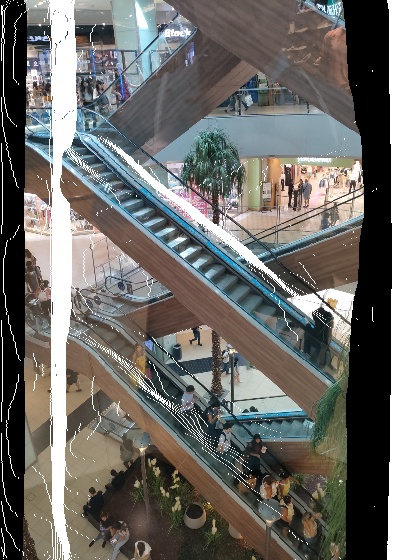}
    \end{subfigure}
    \hfill
    \begin{subfigure}[b]{0.3\textwidth}
        \includegraphics[width=\textwidth]{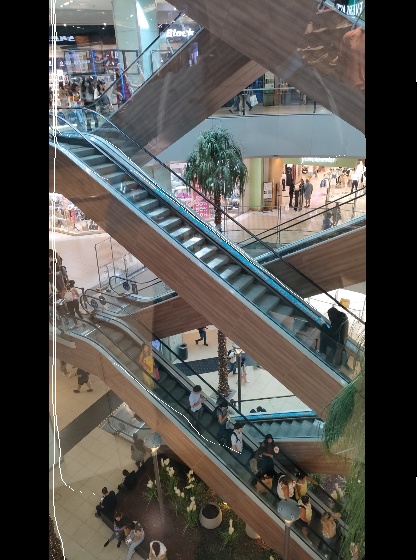} 
    \end{subfigure}
    \caption{A task for image inpainting: Example of holes in some overall virtual digital images.}
    \label{fig:inpainting}
\end{figure}

The CPU time for the computer rendering of a Quilt, and especially the holographic
multiview outputs, varies considerably between different interpolation
algorithms used to obtain a virtual translation motion between consecutive virtual images
and the inpainting techniques adopted.
These include an interpolation for the approximate neighborhood 
pixel intensities, warping, optimization, and the inpainting of occlusions (i.e., empty spots, 
out-of-plane movements as shown e.g. in Fig.~\ref{fig:inpainting}) to get an acceptable illusion of depth and parallax in the horizontal direction.

Inferring depth from a single input image and synthesizing novel views is
a significant challenge because of the huge need for objects information and rich geometric
details (landscape, buildings, sky, etc)~\cite{Eig14,Han22}. In addition to depth ambiguities,
pixels in the generated $N$-views may connect invisible geometries
across regions occluded in the original view causing missing data
that must be handled with inpainting algorithms~\cite{Men20}.
Image inpainting aims to fulfill those missing regions as in the image of Fig.~\ref{fig:inpainting} (in white)
with plausible content and highly depends on the accuracy of the associated depth map.

For inpainting in the generation of multiviews within altiro3D, a default ”Fast” algorithm is
implemented using opencv::remap to map all the pixels
even if the output may become distorted sometimes~\cite{OpenCV}. For the "Real" generation of multiviews, giving as
input an original RGB image and keeping all calibration parameters fixed, altiro3D uses Telea --an standard
image inpainting technique based on the fast marching method~\cite{Tel04}.

\subsection{Neural Network implementation}

The altiro3D library exploits the well-known MiDaS DNN to produce a reasonable depth map from a single image.
The trained network is very effective and can be easily evaluated by using  the DNN module of the OpenCV library. This module can
read the neural network coefficients and inference in an efficient way, the depth map starting from a generic single image.
The format currently used by the altiro3D library is the \verb+onnx+ and the DNN module is configured to prefer CUDA speed-up.
If the hardware is not present, the DNN automatically switches to the CPU.
The depth map can be properly re-scaled to be used in the other stages of the library where views have to be fast generated
to fill-in the Quilt image. From the Quilt image and the LUT, it is then possible to generate the output Native 3D image.

The C++ class devoted to the inference of the neural network inside the altiro3D library is \verb+Network2Quilt+.
This is an application of a generic class \verb+Depth2Quilt+ which implements the optimized reading of the network
coefficients, the DNN module configuration and the creation of the depth map. Currently the class is able to exploit
only the MiDaS 2.1 \verb+model-small.onnx+ and \verb+model-f6b98070.onnx+. The source code will be updated in future releases of altiro3D
to use other updated versions of MiDaS networks. From our test the small network is recommended if the speed is a crucial point.

By default, a "Fast" 
algorithm is adopted here to handle a 3D virtual projection camera and scene transformations along $N$-viewpoints. 
This method analyzes the (down-)degree of depth to move proportionally the pixels, assuming the original 
image to be at the center of all the novel views. This is achieved using the '{\tt cv::remap}' command of 
OpenCV~\cite{OpenCV} by taking pixels from one place in the image and locating them in another position in a new image. 
This fast approach gives reasonable virtual interpretations of reality --at least, within a wide Field of View
 (FoV) --say, $40-100^{o}$. Generating arbitrary number of views can be sometimes cumbersome due to 
occlusion and opening regions leading to in-homogeneous motion fields.

In alternative to our "Fast" method for the generation of multiview images giving as input an original RGB image, 
altiro3D can also be used applying the DIBR algorithm~\cite{Feh04} to synthesize $N$ number of virtual images from 
a (almost equivalent) real camera with $N$ geometric viewpoints between the original camera and the virtual camera.
It requires to calibrate a priori several intrinsic and extrinsic camera for each RGB photo input.

\subsection{Native Image from $N$-Views}

The generation of multiview images starting from a single image (or video frame)
through altiro3D can still offer a potential alternative method for fast 3D vision. Better results 
may be found using MiDaS 3.1 hybrid and large models, but these require extensive computations for 
the conversion and present limits for any real-time 3D streaming. 

\begin{figure}[h]
    \centering
    \includegraphics[width=0.32\textwidth]{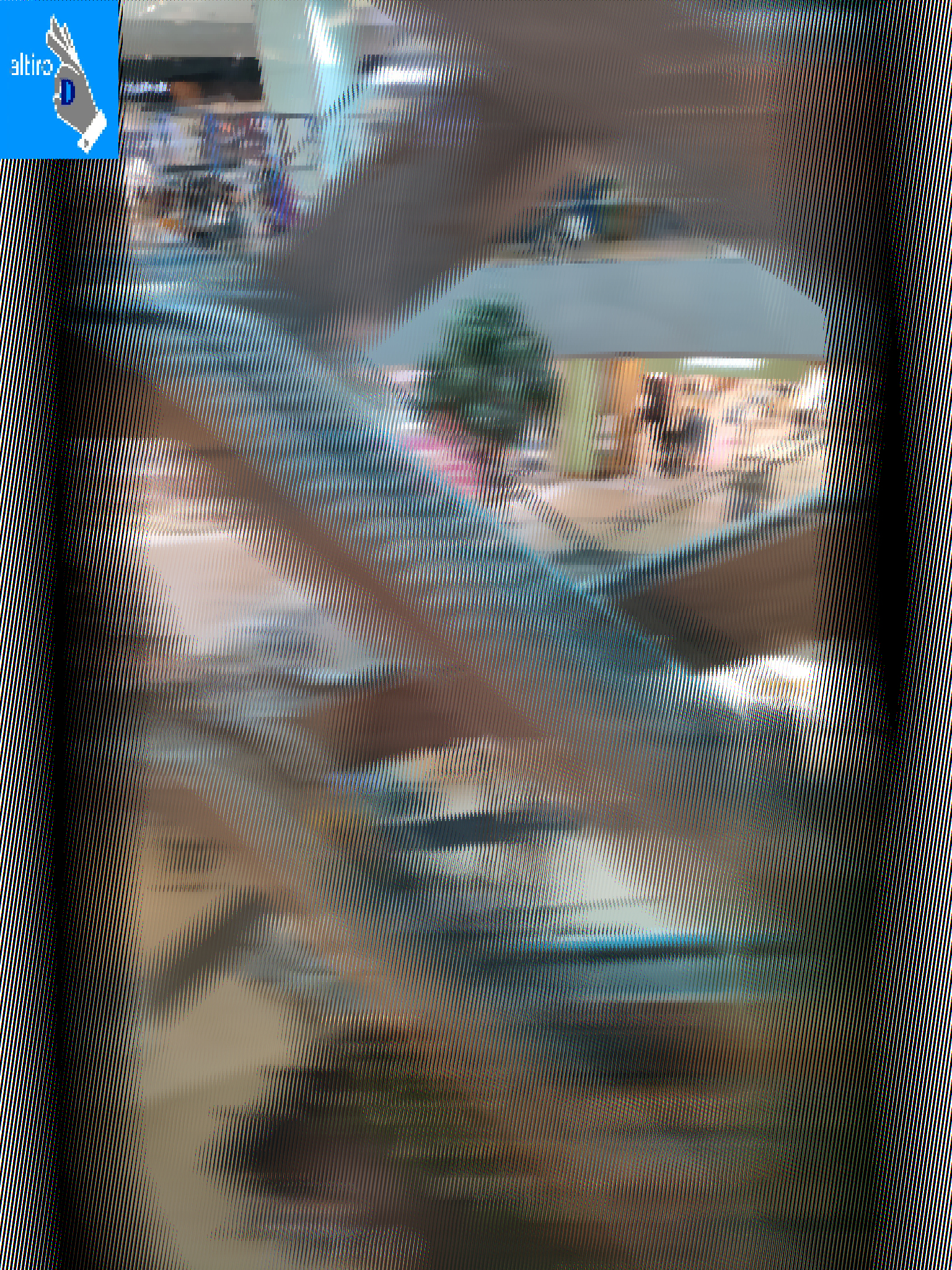}
    \includegraphics[width=0.4\textwidth]{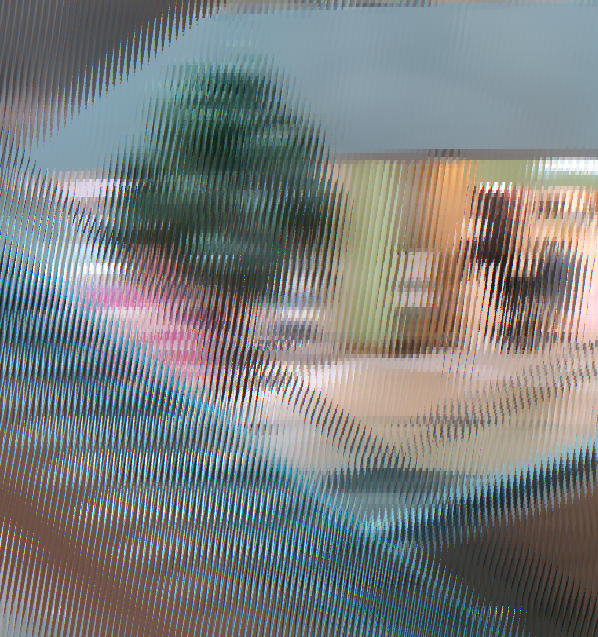}
    \includegraphics[width=0.24\textwidth]{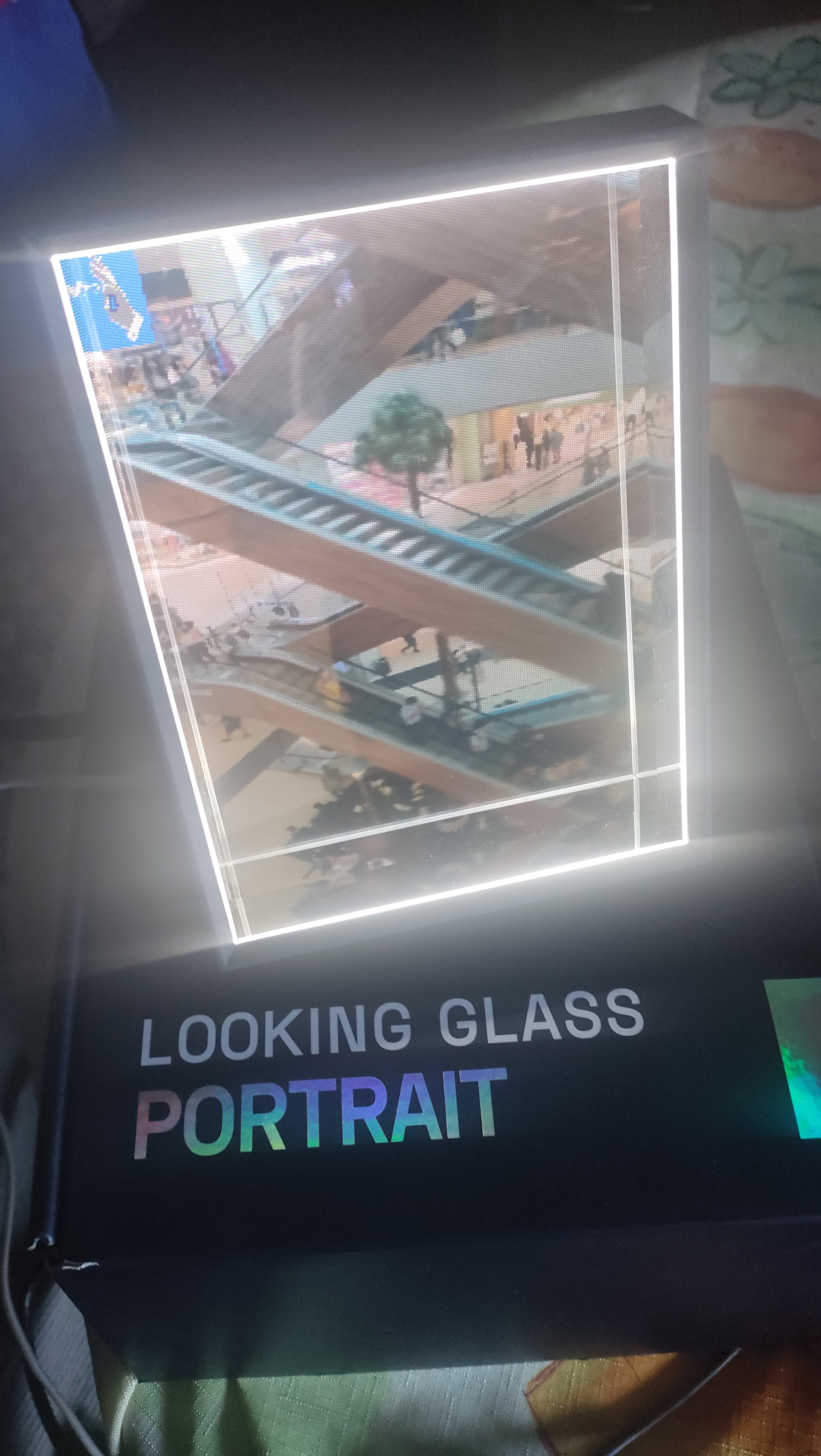} \\
    \caption{Left: Multiview Native output by altiro3D library. Middle: A section of the multiview Native output. Right: 3D display on a headset-free LG Protrait~\cite{lgp} --see also YouTube video: \protect\url{https://www.youtube.com/shorts/hJDVb2TzBr0}}
    \label{fig:multiview}
\end{figure}

With such simple processes involved based on monocular (color or b/w) scene and
novel view synthesis, the altiro3D library provides several different programs.
While the accuracy of the present approach may not be yet competitive with other multiview stereo 
algorithms~\cite{Ana21,Can20,Can20a}, our simpler line of research is particularly promising due 
to the availability of the diverse pre-trained models of MiDaS algorithm.
Using MiDaS 2.1 (small or large) models, altiro3D creates from photo or videos frames a 3D Native 
image or a 3D video, respectively. The altiro3D library also allows to create  
Native from still $N$-views, i.e., sorted in sequential order and stored in a given directory, with
output as in Fig.~\ref{fig:multiview}, and convert a given 2D video to Native 3D video (.mp4) as well.

Scaling $N\times M$ views --e.g., scaling from the $(6\times 8)$ Quilt (of size $3360\times 3360 px$) to a Native
image (of size $1536\times 2048 px$) with an output illustrated as in Fig.~\ref{fig:multiview}, leads to increase system complexity
and requires lots of CPU resources. However, as we discussed in~\cite{Can20,Can20a},
this problem can be reduced with the generation of a device-dependent LUT table. 

The class concerning the views creation is \verb+Depth2Quilt+ implemented starting from the evaluated 
depth map and from the input image view. As anticipated, the current class provides two rendering methods: "Fast" and "Real".
The "Fast" method is an efficient method to fast create multiple views to feed the Quilt image. The "Fast" implementation 
leverages on the fast remap method provided by the OpenCV library~\cite{OpenCV}. This is a simplified method and does not 
use a real model based on intrinsic and extrinsic matrices of camera. There are simple approaches to estimate
the focal length of a camera given the horizontal FoV and image width~\cite{Wor15}.

altiro3D simply creates a dense map to the new pose from the original 
image which is considered as the central image (mapped in the central position of the Quilt) 
and the depth as proportional to the distance. In a dual-camera depth reconstruction algorithm, the depth map is related 
to the difference position of the same object in the reference frames of two cameras, so called "disparity map". So the 
depth can be interpreted as produced by a disparity map and it can be used to remap pixels of the central image according 
to the depth/disparity map. The process used in our simpler "Fast" method maps pixels in the new pose accordingly and proportionally 
to the depth. Then the produced map is applied to the original by the OpenCV remap function which is implemented with a 
linear interpolation algorithm. Such interpolation algorithm avoids artifacts due to occlusion and dis-occlusion with 
a modest image deformation.
The flow of the "Fast" algorithm is shown with pseudo-code in Algorithm~\ref{alg:fast}.

\begin{algorithm}[H]
\caption{Fast algorithm to create intermediate views}
\label{alg:fast}
\begin{algorithmic}
\Require depthMap, offset, cols, rows, origImage
\Ensure destImage
\Function{GetFastMapFromDepth}{depthMap, offset, cols, rows}
\State MapX $\gets 0$
\State MapY $\gets 0$
\For{c = 1 to cols}
    \For{r = 1 to rows}
        \State MapX[r,c]$ = r$
        \State MapY[r,c]$ = c - $depthMap[r,c]$ * $offset
    \EndFor
\EndFor
\State \textbf{return} MapX, MapY
\EndFunction 
\\
\State MapX$, $MapY $\gets$ \Call{GetFastMapFromDepth}{depthMap, offset, cols, rows}
\State destImage $\gets$ \Call{cv::remap}{origImage, MapX, MapY}
\end{algorithmic}
\end{algorithm}

Where the \textbf{depthMap} is the depthmap obtained by the MiDaS network, \textbf{offset} is a multiplicative constant controlling the output view position, \textbf{cols} and \textbf{rows} are the the width and the height of the input image, \textbf{origImage} is the image acquired by the camera.
If ``specular views'' have to be produced, the function \textbf{GetFastMapFromDepth} has to be called twice: the first time with \textbf{offset} and the second time with \textbf{-offset}.

On the other hand, the "Real" method considered is a more sophisticated implementation of a real camera 
DIBR model~\cite{Feh04}. It takes into account two matrices, intrinsic and extrinsic, to produce the new real pose. 
The intrinsic matrix is related to the acquisition parameters such as focal and center of camera frame, while the 
extrinsic matrix is related to the position and attitude of the camera frame relative to the real world frame, 
where the original object is located. By modifying the extrinsic camera it is possible to create a new view of 
the original image: the extrinsic matrix allows rotation or translation of the frame.
This method takes into account the real model to get new views, but it produces more artifacts due to the occlusion 
and dis-occlusion effect. In order to reduce artifacts in the final view (c.f., Fig.~\ref{fig:inpainting}), we apply an inpainting 
technique~\cite{Tel04}, along with a median spatial filter. 

\begin{algorithm}[H]
    \caption{Real algorithm to create intermediate views}\label{alg:real}
    \begin{algorithmic}
        \Require depthMap, offset, origImage
        \Ensure dst1, dst2
        \Function{GetRealMapFromDepth}{origImage, depthMap, offset, K$_o$, R$_o$, K$_v$, R$_v$}
            \State R $\gets$ init rotation matrix
            \State T $\gets$ init translation vector
            \State T[0] $\gets$ offset
            \State $\text{R}_v$ $\gets$ createExtrinsicMatrix(R, T)
            \State \Call{setMatrices}{K$_o$, R$_o$, K$_v$, R$_v$} \Comment{To create transform from original matrices to the new virtual pose}
            \State \Call{setInput}{origImage, depthMap} \Comment{To set input image and depth}
            \State \Call{evalTransform}{} \Comment{To create a new view}
            \State \textbf{return} \Call{getProducedView}{}
        \EndFunction
        \\
        \For{i = 1 to numberOfViews/2} \Comment{Main loop}
            \State offset $\gets$ position of view to be produced
            \State dst1 $\gets$ \Call{GetRealMapFromDepth}{origImage, depthMap, offset, K$_o$, R$_o$, K$_v$, R$_v$}
            \State dst2 $\gets$ \Call{GetRealMapFromDepth}{origImage, depthMap, -offset, K$_o$, R$_o$, K$_v$, R$_v$}
        \EndFor
    \end{algorithmic}
\end{algorithm}

The "Real" method may be more effective, but the evaluation of view is computational intensive
compared to the alternative "Fast" method, so it is not well suited for real-time implementations.
To obtain geometric viewpoints between the original and virtual camera, the "Real" method
also require calculation power for generating a final Native image with resolution $3360\times 3360 px$.
RGB color channels for a $6\times 8$ Quilt --such that, once the pixel to be mapped is fixed, the map
value for each color channel implies separated calculations. In essence this procedure as such makes real-time
video in 3D difficult to achieve.

The pseudo-code of the "Real" algorithm is summarized in Algorithm~\ref{alg:real}.
In main loop the views are created according to the value of the \textbf{offset} parameter. Every loop produces two output images with a symmetric \textbf{offset} value. The function {\sc GetRealMapFromDepth} sets all the needed structures to evaluate the transform from the single image to the required view. This function exploits a camera model composed by intrinsic and extrinsic matrix: the function generates the transformation from the original set of matrices (K$_o$, R$_o$) to the virtual pose (K$_v$, R$_v$).

\section{\label{sec:conclusion}Conclusion and future work}

We introduced the altiro3D C++ library to synthesize {\it N}-number of virtual images and add 
them sequentially into a Quilt collage by applying MiDaS code for the monocular depth estimation. 
Novel view synthesis from a single image is carried out by
using simple inpainting techniques to map all pixels, and implementing "Fast" and "Real"
algorithms for the camera and scene transformations along {\it N}-viewpoints.
A unique pixel- and device-based LUT to optimize computing time is implemented.
In the absence of a LUT procedure, it would become computationally expensive and difficult 
to apply many techniques for real-time video in 3D.

This latter aspect of our algorithm could stimulate further investigations toward real-time 3D applications 
deployed from Desktop computers and/to mobile devices~\cite{Hac20}. Streaming in real time a hologram feed is
computationally demanding, because of the larger amount of information contained in the many light-fields 
to be streamed live, as compared to sending "realistic" frames (of monocular views) via 2D video streams. 
Finally, removing the dependency on stereoscopic images as input ~\cite{Can20,Can20a}, it makes our altiro3D algorithm more
widely applicable to a larger amount of entire (historical) datasets. The generated images may be further 
improved by obtaining more "realistic" perspectives from recent machine learning or deep learning algorithms
of MiDaS 3.1 to obtain meaningful information~\cite{Ran22}. 

In future work, we want to extend the present scene static representation from a single image to 
a more dynamical, glasses free live 3D vision. The $N$-view synthesis of altiro3D in this case 
requires a fast conversion of each video frame into native light-field images, and respective Quilts, 
at a reasonable frame rate of at least $10fps$ in order to get a free-viewpoint real-time streaming 
on lenticular displays. These interesting directions have tremendous potential to be explored.

\section{\label{sec:sources}Source}

Further information, binaries, papers, presentations, manuals, or to report bugs, can be found at \\
\url{https://github.com/canessae/altiro3D}

\section{\label{sec:interest}Conflict of interest}

The authors have no competing interests to declare that are relevant to the content of this article.

\newpage

\section*{\label{sec:graph}Graphical abstract}

\begin{figure}[ht]
    \centering
        \includegraphics[width=\textwidth]{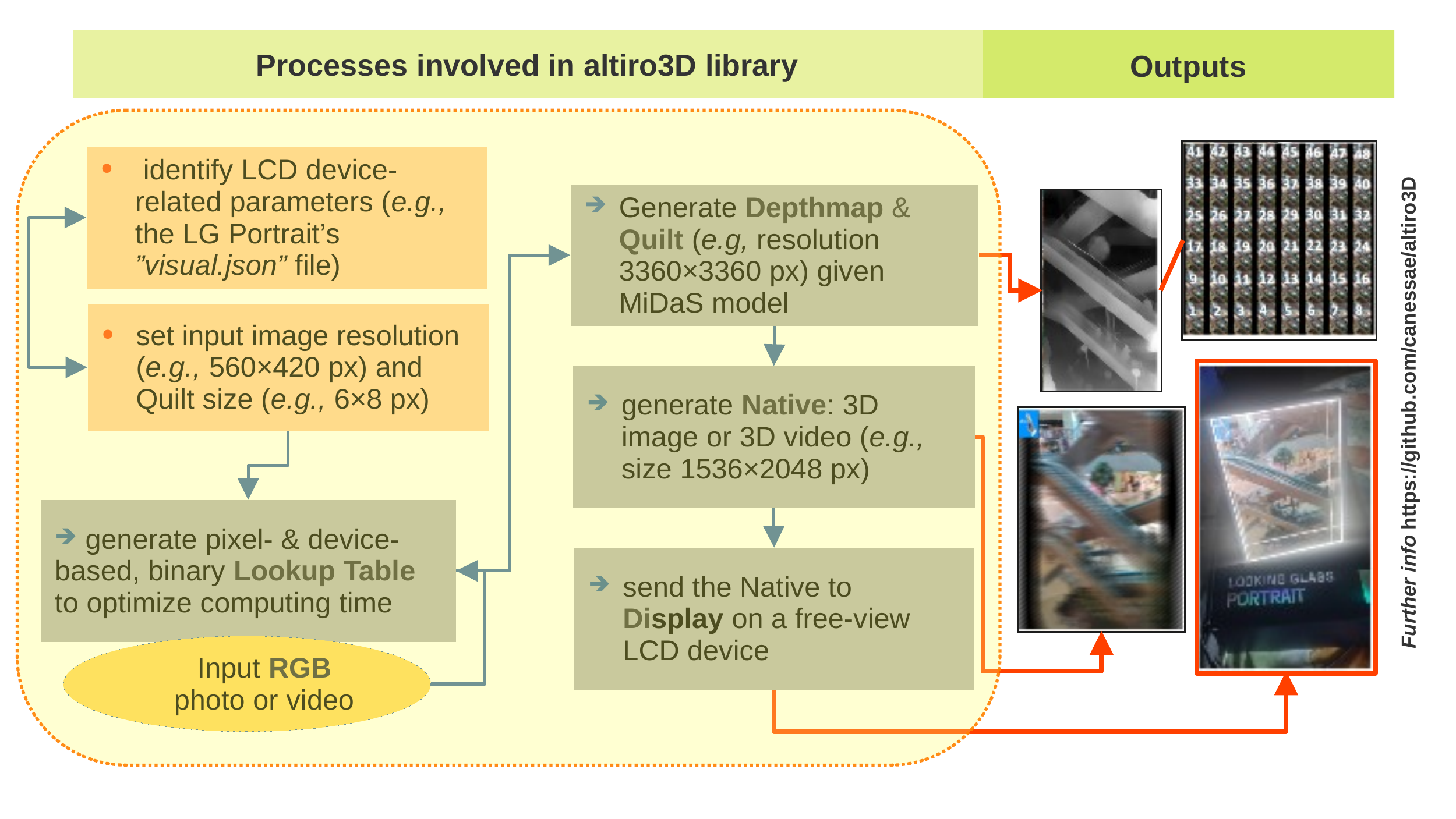}
    \caption{Flow diagram of altiro3D.}
    \label{fig:visualabstract}
\end{figure}


\begin{thebibliography}{10}

\bibitem{Ana21}
Anantrasirichai N., Geravand M. et.al.,
"Fast Depth Estimation for View Synthesis",
28th European Signal Processing Conference (EUSIPCO) 2021, 575-579.
doi: 10.23919/Eusipco47968.2020.9287371
ArXiv: \url{https://arxiv.org/abs/2003.06637} (Last visited 30/3/2023)
doi: 10.48550/arXiv.2003.06637

\bibitem{Can20}
Canessa E., Tenze L., 
"Morpholo: A Hologram Generator Algorithm",
Electronic Imaging {\bf 53} (2020) 53-1-53-5.
doi: 10.2352/ISSN.2470-1173.2020.2.SDA-053

\bibitem{Can20a}
Canessa E., Tenze L., 
"Morphing a Stereogram into Hologram",
J. Imaging {\bf 6} (2000) 1.
doi: 10.3390/jimaging6010001 

\bibitem{Eig14}
Eigen D., Puhrsch C. et al.,
"Depth Map Prediction from a Single Image using a Multi-Scale Deep Network",
NIPS'14: Proc. 27th Intl. Conf. Neural Information Process. Sys. {\bf 2} (2014) 2366–2374.
doi: 10.5555/2969033.2969091 
ArXiv: \url{https://arxiv.org/abs/1406.2283} (Last visited 30/3/2023)
doi: 10.48550/arXiv.1406.2283

\bibitem{Cha20}
Zhao Ch., Sun Q.Y. et al.,
"Monocular depth estimation based on deep learning: An overview",
Sci China Tech Sciences {\bf 63} (2020) 1612–1627.
doi: 10.1007/s11431-020-1582-8
ArXiv: \url{https://arxiv.org/abs/2003.06620} (Last visited 30/3/2023)
doi: 10.48550/arXiv.2003.06620

\bibitem{Min21}
Ming Y., Meng X. et al.,
"Deep learning for monocular depth estimation: A review",
Neurocomputing {\bf 438} (2021) 14–33.
doi: 10.1016/j.neucom.2020.12.089

\bibitem{Ran22}
Ranftl R., Lasinger K. et al.,
"Towards Robust Monocular Depth Estimation: Mixing Datasets for Zero-shot Cross-dataset Transfer",
IEEE Trans. Pattern Analysis. Mach. Intell. {\bf 44} (2022) 1623-1637.
doi: 10.1109/TPAMI.2020.3019967
ArXiv: \url{https://arxiv.org/abs/1907.01341v3} (Last visited 30/3/2023)
doi: 10.48550/arXiv.1907.01341

\bibitem{Zoe23}
Bhat S.F., Birkl R. et al.,
"ZoeDepth: Zero-shot Transfer by Combining Relative and Metric Depth",
Arxiv: \url{https://arxiv.org/abs/2302.12288} (Last visited 30/3/2023)
doi: 10.48550/ARXIV.2302.12288

\bibitem{Pan23}
Pandey J., Asati A.R., 
"Lightweight convolutional neural network architecture implementation using TensorFlow lite",
Int. j. inf. tecnol. {\bf 15} (2023) 2489–2498. doi: 10.1007/s41870-023-01320-9

\bibitem{Chau23}
Chaurasiya R., Ganotra D.,
"Deep dilated CNN based image denoising",
Int. j. inf. tecnol. {\bf 15} (2023) 137–148. doi: 10.1007/s41870-022-01125-2

\bibitem{OpenCV}
OpenCV --simple remapping and impainting implementation:
\url{https://docs.opencv.org/3.4/d1/da0/tutorial\_remap.html} and
\url{https://docs.opencv.org/3.4/df/d3d/tutorial\_py\_inpainting.html} (Last visited 30/3/2023)

\bibitem{Tel04}
Telea A.,
"An image inpainting technique based on the fast marching method",
J. Graphics Tools, {\bf 9} (2004) 23-34.
doi: 10.1080/10867651.2004.10487596

\bibitem{Feh04}
Fehn C.,
"Depth-image-based rendering (DIBR), compression, and transmission for a new approach on 3D-TV",
Proc. SPIE 5291, Stereoscopic Displays and Virtual Reality Systems XI (2004).
doi: 10.1117/12.524762
Code available at \url{https://github.com/3ZadeSSG/DIBR-Algorithm} (Last visited 30/3/2023)

\bibitem{Tak11}
Takaki Y., Tanaka K., Nakamura J.,
"Super multi-view display with a lower resolution flat-panel display",
Optics Express {\bf 19} (2011) 4129.
doi: 10.1364/OE.19.004129 

\bibitem{lgp} 
Low cost Looking Glass Portrait: \url{https://lookingglassfactory.com/looking-glass-portrait}
(Last visited 30/3/2023)

\bibitem{Wan23}
Wang J., Chen Y. et al.,
"SABV-Depth: A biologically inspired deep learning network for monocular depth estimation",
Knowledge-Based Systems {\bf 263} (2023) 110301-14
doi: 10.1016/j.knosys.2023.110301

\bibitem{Sha23}
Shamalik R., Koli S.,
“FabDepth I: A Unique Dataset for Efficient Gesture Detection”,
Int. j. inf. tecnol. {\bf 15} (2023) 2645–2649. 
doi: 10.1007/s41870-023-01295-7

\bibitem{Che22}
Chetty G., Yamin M., White M.,
"A low resource 3D U-Net based deep learning model for medical image analysis",
Int. j. inf. tecnol. {\bf 14} (2022) 95–103. doi: 10.1007/s41870-021-00850-4

\bibitem{Cha23}
Chaurasia R.K., Jaiswal U.C.,
"Spatio-temporal based video anomaly detection using deep neural networks",
Int. j. inf. tecnol. {\bf 15} (2023) 1569–1581. doi: 10.1007/s41870-023-01193-y

\bibitem{QT}
QT ("cute") software to create graphical user interfaces ans cross-platform applications.
\url{https://www.qt.io/}

\bibitem{OpenCV2}
Open Source Computer Vision Library (OpenCV): an open source computer vision and machine learning software library.
\url{https://opencv.org/}

\bibitem{Doxygen}
Doxygen de facto standard tool for generating documentation from annotated C++ sources.
\url{https://www.doxygen.nl/}

\bibitem{Ber97}
van Berkel C., Clarke J.A., 
“Characterization and optimization of 3D-LCD module design”,
Proc. SPIE {\bf 3012} (1997) 179–186.
doi: 10.1117/12.274456

\bibitem{Han22}
Han Y., Wang R. et al.,
"Single-View View Synthesis in the Wild with Learned Adaptive Multiplane Images",
Proc. ACM SIGGRAPH Article 14 (2022) 1–8.
doi: 10.1145/3528233.3530755

\bibitem{Men20}
Meng-Li S., Shih-Yang S. et al.,
"3D Photography using Context-aware Layered Depth Inpainting",
IEEE Conf. Comp. Vision and Pattern Recognition (CVPR) 2020
ArXiv: \url{https://arxiv.org/abs/2004.04727} (Last visited 30/3/2023)
doi: 10.48550/arXiv.2004.04727

\bibitem{Wor15}
Workman S., Greenwell C.,
"DEEPFOCAL: A Method for Direct Focal Lenght Estimation".
Proc. IEEE International Conference on Image Processing -ICIP (2015).
doi: 10.1109/ICIP.2015.7351024

\bibitem{Hac20}
DIY Arduino Parallax 3D Display: \url{https://hackaday.io/project/174756-diy-arduino-parallax-3d-display}
(Last visited 30/3/2023)

\end{thebibliography}
\end{document}